\documentclass{salt}

\usepackage[equation]{gb4e-salt}
\noautomath

\usepackage{fdsymbol}
\usepackage{amsmath,amssymb,mathtools,bm}
\usepackage{xspace}
\usepackage{linguex}
\usepackage{tabularx}
\usepackage{todonotes}

\pdfauthor{Gene Louis Kim, Aaron Steven White}
\pdftitle{Montague Grammar Induction}
\pdfkeywords{grammar induction, combinatory categorial grammar, semantic selection, experimental semantics, computational semantics, experimental syntax, computational syntax}

\title[Montague Grammar Induction]{Montague Grammar Induction%
  \thanks{We are grateful to the audience at SALT 30 for discussion of this work. Funding was provided by NSF-BCS-1748969 (\textit{The MegaAttitude Project: Investigating selection and polysemy at the scale of the lexicon}) and NSF EAGER grant IIS-1908595 (\textit{Learning a High-Fidelity Semantic Parser}).
}
}

\author[Kim \& White]{%
  \saltauthor{Gene Louis Kim \\ \institute{University of Rochester}} \AND
  \saltauthor{Aaron Steven White \\ \institute{University of Rochester}}%
}

\newcommand{\denotation}[1]{\llbracket #1 \rrbracket}

\newcommand{\vocabulary}{\mathcal{V}_\text{lex}}
\newcommand{\parserspace}{\mathcal{V}_\text{node}}
\newcommand{\interpretationspace}{\mathcal{V}_\text{interp}}

\newcommand{\prob}{\mathbb{P}}
\newcommand{\reals}{\mathbb{R}}

\newcommand{\types}{\mathcal{T}}
\newcommand{\combinators}{\mathcal{C}}
\newcommand{\directions}{\mathcal{D}}
\newcommand{\primitivetypes}{\mathcal{P}}
\newcommand{\loss}{\mathcal{L}}
\newcommand{\actions}{\mathcal{A}}

\newcommand{\entity}{\text{\sffamily e}}
\newcommand{\state}{\text{\sffamily s}}
\newcommand{\truthvalue}{\text{\sffamily t}}

\newcommand{\montaguetype}[2]{\langle #1, #2 \rangle}

\newcommand{\Wb}{\mathbf{W}}
\newcommand{\Xb}{\mathbf{X}}
\newcommand{\xb}{\mathbf{x}}
\newcommand{\hb}{\mathbf{h}}

\newcommand{\vb}{\mathbf{v}}

\newcommand{\hinside}[1]{\hb^{\medtriangleup}_{#1}}
\newcommand{\houtside}[1]{\hb^{\medtriangledown}_{#1}}
\newcommand{\hinsideoutside}[1]{\hb^{\diamond}_{#1}}

\newcommand{\et}{\montaguetype{\entity}{\truthvalue}}
\newcommand{\st}{\montaguetype{\state}{\truthvalue}}
\newcommand{\sst}{\montaguetype{\state}{\st}}
\newcommand{\est}{\montaguetype{\entity}{\st}}
\newcommand{\eet}{\montaguetype{\entity}{\et}}
\newcommand{\eest}{\montaguetype{\entity}{\est}}
\newcommand{\sest}{\montaguetype{\state}{\est}}
\newcommand{\ett}{\montaguetype{\entity}{\montaguetype{\truthvalue}{\truthvalue}}}

\newcommand{\eett}{\montaguetype{\entity}{\ett}}
\newcommand{\eeet}{\montaguetype{\entity}{\eet}}

\xspaceaddexceptions{]\}}
\newcommand{\actionfn}[1]{\textsc{action}$_\text{#1}$\xspace}
\newcommand{\primitivefn}[1]{\textsc{primitive}$_\text{#1}$\xspace}

\newcommand{\attendfn}[1]{\textsc{attend}$_\text{#1}$\xspace}

\newcommand{\raisefn}[1]{\textsc{raise}$_\text{#1}$\xspace}

\begin{document}


\maketitle

%
\setcounter{page}{1}

\firstpageheadings{30}{000}{000}{2020}{}{}

\begin{abstract}  
  We propose a computational modeling framework for inducing combinatory categorial grammars from arbitrary behavioral data. This framework provides the analyst fine-grained control over the assumptions that the induced grammar should conform to: (i) what the primitive types are; (ii) how complex types are constructed; (iii) what set of combinators can be used to combine types; and (iv) whether (and to what) the types of some lexical items should be fixed.  In a proof-of-concept experiment, we deploy our framework for use in distributional analysis. We focus on the relationship between s(emantic)-selection and c(ategory)-selection, using as input a lexicon-scale acceptability judgment dataset focused on English verbs' syntactic distribution (the \href{http://megaattitude.io/projects/mega-acceptability/}{MegaAcceptability} dataset) and enforcing standard assumptions from the semantics literature on the induced grammar.

\end{abstract}



\setlength{\Exlabelsep}{.3em}
\setlength{\Extopsep}{.6\baselineskip}

\setlength{\SubExleftmargin}{1.5em}

\section{Introduction}
\label{sec:introduction}

Semantic theories aim to capture two kinds of facts about languages' expressions: (i) their distributional characteristics; and (ii) their inferential affordances. The descriptive adequacy of any such theory is evaluated in terms of its coverage of these facts---an evaluation that is commonly carried out informally on a relatively small number of test cases. While this approach to theory-building and evaluation has yielded deep insights, it also carries significant risks: generalizations that appear unassailable based a small number of high-frequency examples (and theories built on them) can collapse when evaluated on a more diverse range of expressions purportedly covered by the generalization (see \citealt{white_believing_accepted} for recent discussion). 

Ameliorating these risks requires developing scalable methods for building and evaluating semantic theories. Until recently, a major obstacle to developing such methods was that sufficiently large-scale behavioral datasets were not available. This situation has changed with the advent of lexicon-scale acceptability and inference judgment datasets, such as the \href{http://megaattitude.io}{MegaAttitude} datasets \citep{white_computational_2016,white_role_2018,white_frequency_accepted,white_lexicosyntactic_2018,an_lexical_2020,moon_source_toappear}. Concomitant advances in computational modeling have cleared the way for automating distributional and inferential analysis for the purposes of theory-building and evaluation. The remaining challenge is one of integration. On the one hand, powerful methods for learning structured representations from corpus data and (to some extent) behavioral data now exist \citep{le_inside-outside_2014,le_compositional_2015,williams_latent_2018,shen_neural_2018,kim_unsupervised_2019,drozdov_unsupervised_labeled_2019,drozdov_unsupervised_latent_2019}, but the relationship between these models' representations and grammars posited under standard frameworks (\citealt{montague_proper_1973} \textit{et seq}) assuming some form of (combinatory) categorial grammar \citep{steedman_syntactic_2000} remains unclear.\footnote{Other systems \textit{assume} prior knowledge of some, often quite exquisite, structure (see \citealt{baroni_frege_2014} and references therein). We focus here on systems that \textit{learn} said structure.} On the other hand, powerful methods for inducing such grammars now exist \citep{zettlemoyer_learning_2005,zettlemoyer_online_2007,zettlemoyer_learning_2009,kwiatkowksi_inducing_2010,kwiatkowski_lexical_2011,bisk_induction_2012,bisk_simple_2012,bisk_hdp_2013,bisk_probing_2015}, but they do not straightforwardly generalize to the full range of behavioral data of interest in experimental semantics.

To address these limitations, we propose a general deep learning-based, computational modeling framework for inducing full-fledged combinatory categorial grammars from multiple distinct types of behavioral data. Beyond providing the ability to synthesize arbitrary distributional and inferential data within a single model, our framework provides the analyst fine-grained control over the assumptions that the induced grammar should conform to: (i) what the primitive types are (e.g. $\entity$, $\state$, $\truthvalue$, etc.); (ii) how complex types are constructed (e.g. that $\montaguetype{t_1}{t_2}$ is a type if $t_1$ and $t_2$ are types); (iii) what set of combinators can be used to combine types (e.g. application, composition, etc.); and (iv) whether (and to what) the types of some lexical items should be fixed.  As a proof of concept, we deploy our framework for use in distributional analysis. We focus, in particular, on the relationship between s(emantic)-selection and c(ategory)-selection, using as input a lexicon-scale acceptability judgment dataset focused on English verbs' syntactic distribution (the \href{http://megaattitude.io/projects/mega-acceptability/}{MegaAcceptability} dataset; \citealt{white_computational_2016,white_frequency_accepted}) and enforcing standard assumptions from the semantics literature on the induced grammar. As a case study, we analyze the typing that the induced grammar infers for clausal complements. Clausal complements are useful in this regard, since their syntactic complexity provides a rigorous test of our framework's ability to recover interpretable types.

We begin with some brief background on the deep learning-based approach to grammar induction that we build on (\S\ref{sec:background}). We then describe how we extend that approach to learn full-fledge combinatory categorial grammars (\S\ref{sec:computational-model}) before turning to our proof-of-concept experiments with the MegaAcceptability dataset (\S\ref{sec:experiments}) and the results of these experiments (\S\ref{sec:results}). We conclude with a discussion of future directions for our framework (\S\ref{sec:conclusion}).

\section{Deep learning-based approaches to grammar induction}
\label{sec:background}

Our framework fits within the broader context of \textit{grammar induction} (or \textit{grammatical inference}; see \citealt{heinz_topics_2016} for recent reviews)---in particular, distributional learning-based approaches (see \citealt{clark_distributional_2016} for a recent review). The goal of a grammar induction system is to produce a grammar for some language based on a dataset containing (at least) well-formed expressions of that language. These datasets may simply be (multi)sets or sequences of such expressions or they may further associate labels with expressions or sequences thereof---e.g. associating a (possibly malformed) expression with its acceptability or a pair of expressions with a label indicating whether the first entails the second. We focus specifically on \textit{supervised} grammar induction, wherein some labeling of expressions is assumed, since most behavioral datasets fit this description.

The grammars (or weightings thereon) output by such a system can take a variety of forms. Here, we build on grammar induction systems that learn to encode both expressions and grammars in a vector space, treating the problem as one of training the parameters of a machine learning model to predict the labels in some dataset---e.g. to predict acceptability or entailment. A major benefit of using such a system is that it is straightforward to train on arbitrary behavioral data (see \citealt{potts_case_2019} for further discussion): given a way of mapping expressions to vectors, those vectors can be input to standard regression models from the experimental literature and off-the-shelf optimization routines employed to jointly train the grammar induction system and regression model (see \citealt{goldberg_neural_2017} for a technical review).

We specifically build on recently developed methods that generalize the \textit{inside-outside algorithm} \citep{baker_trainable_1979} for training probabilistic syntactic parsers to vector space syntactic parsers. These methods recursively construct vector representations of expressions aimed at capturing their syntactic features and/or semantic properties. These methods have two interlocking components: (i) a generalization of the \textit{inside algorithm}, which constructs a vector representation for an expression based on the vector representations of its subexpressions; and (ii) a generalization of the \textit{outside algorithm}, which constructs a vector respresentation for an expression based on the vector representations of surrounding expressions \citep{le_inside-outside_2014,le_compositional_2015,drozdov_unsupervised_labeled_2019,drozdov_unsupervised_latent_2019}. The former can be thought of very roughly as implementing a parameterized version of Minimalism's \textsc{merge} (or similar ``bottom-up'' operations); and the latter can be thought of (again, very roughly) as implementing a parameterized version of Minimalism's \textsc{agree} (or similar ``top-down'' operations).

Both the inside and outside algorithms compute probabilities relative to a string and a probabilistic context free grammar (see \citealt{manning_foundations_1999} for an in-depth discussion). But for understanding the relevance of these algorithms here, it is possible to ignore the probabilistic aspects, starting by assuming a vanilla context free grammar (CFG) in Chomsky Normal Form---i.e. with rules of the form $A \rightarrow B\;C$ or $A \rightarrow w$, where $A$, $B$, and $C$ nonterminals and $w$ a terminal---and then generalizing. 

In this non-probabilistic context, the inside algorithm is analogous to the CKY algorithm for recognition and parsing \citep{younger_recognition_1967}: given a grammar $\mathcal{G}$ and a sentence $S$ consisting of a sequence of words $w_0...w_{|S|-1}$, the algorithm computes, for each subsequence of words $w_{i:j} = w_i...w_{j-1}$ in $S$ (where $0 \leq i < j \leq |S|$), the set $N_{ij}$ of nonterminals that could yield $w_{i:j}$---i.e. the nonterminals for which some sequence of rewrites would result in $w_{i:j}$. These sets are computed recursively: let $N_{i(i+1)} \equiv \text{\sc unary}_\mathcal{G}(w_i)$ for $0 \leq i < |S|$ and $N_{ij} \equiv \bigcup_{k=i+1}^{j-1} \text{\sc combine}_\mathcal{G}(N_{ik}, N_{kj})$ for $0 \leq i < j-1 < |S|$, where $\text{\sc unary}_\mathcal{G}(w) = \{A \;|\; (A \rightarrow w) \in \mathcal{G}\}$ and $\text{\sc combine}_\mathcal{G}(N, N') = \bigcup_{\langle B, C \rangle \in N \times N'} \{A \;|\; (A \rightarrow B\;C) \in \mathcal{G}\}$ \citep[see][]{shieber_principles_1995}.

Defined in this way, $N_{ij}$ specifies what kinds of constituent $w_{i:j}$ could be (e.g. a noun phrase, a verb phrase, etc.), ignoring the rest of the sentence. The non-probabilistic analogue of the outside algorithm computes $O_{ij}$: what kinds of constituent $w_{i:j}$ could be considering only the parts of the sentence outside $w_{i:j}$. The intersection $N_{ij} \cap O_{ij}$ then indicates all and only the kinds of constituents $w_{i:j}$ could be in the context of the entire sentence. Like $N_{ij}$, $O_{ij}$ is computed recursively: assume a fixed $O_{0|S|}$ and let $O_{ij} \equiv \bigcup_{m=j+1}^{|S|} \text{\sc split}^R_\mathcal{G}(O_{im}, N_{jm}) \cup \bigcup_{m=0}^{i-1}\text{\sc split}^L_\mathcal{G}\left(O_{mj}, N_{mi}\right)$ for $0 \leq i < j-1 < |S|$, where $\text{\sc split}^R_\mathcal{G}(O, N) = \bigcup_{\langle A, C \rangle \in O \times N} \{B \;|\; (A \rightarrow B\;C) \in \mathcal{G}\}$ and $\text{\sc split}^L_\mathcal{G}(O, N) = \bigcup_{\langle A, B \rangle \in O \times N} \{C \;|\; (A \rightarrow B\;C) \in \mathcal{G}\}$.

Importantly for our purposes, this algorithm does not require the grammar to be a CFG or even a probabilistic generalization thereof. For example, with straightforward modifications to \textsc{unary}, \textsc{combine}, \textsc{split}$^L$, and \textsc{split}$^R$, it can be adapted to arbitrary combinatory categorial grammars (CCGs) and therefore the sorts of type grammars familiar in semantic theory. \textsc{unary} returns the set of types listed for (the denotation of) a word; and given some set of combinators $\combinators$---e.g. \textsc{application}, \textsc{composition}, etc.---$\text{\sc combine}_G(N, N') = \bigcup_{\langle t, t' \rangle \in N \times N'} \{c(t, t') \;|\; c \in \combinators \land \langle t, t' \rangle \in \text{dom}(c)\}$ (with \textsc{split}$^L$ and \textsc{split}$^R$ analogously defined). This possibility becomes important for our framework, discussed in detail in \S\ref{sec:computational-model}.

The generalization of these algorithms to vector spaces involves four main changes: (i) substituting terminals with vectors in $\vocabulary = \reals^{M_\text{lex}}$ and sets $N_{ij}, O_{ij}$ with vectors in $\parserspace = \reals^{M_\text{node}}$; (ii) substituting the function \textsc{unary} with a parameterized function from $\vocabulary$ to $\parserspace$ and the functions \textsc{combine}, \textsc{split}$^R$, and \textsc{split}$^L$ with parameterized functions from $\parserspace^2$ to $\parserspace$; (iii)  substituting unions with weighted vector sums; and (iv) substituting intersections and products with vector concatenations.\footnote{Unless otherwise specified, we implement parameterized functions as multi-layer perceptrons with a single LeakyReLU hidden layer of the same size as the output~\citep{maas2013rectifier}: $\text{MLP}(\xb) = \Wb_1 \text{LeakyReLU} (\Wb_2 \xb + \mathbf{b}_2) + \mathbf{b}_1$, where the $\Wb$s and $\mathbf{b}$s are the parameters to be optimized.} The analogue to the symbolic grammar $\mathcal{G}$ is then the mapping from terminals to vectors and the parameterization of these functions. 

More specifically, we define \textit{inside vectors} $\hinside{ij}$, analogous to $N_{ij}$, and \textit{outside vectors} $\houtside{ij}$, analogous to $O_{ij}$. Assuming some mapping from $w_0...w_{|S|-1}$ to a sequence of vectors $\xb_0...\xb_{|S|-1}$ and that $\hinside{i(i+1)} = \text{\sc unary}(\xb_i)$ for $0 \leq i < |S|$:

\vspace{-5mm}

\[\hinside{ij} = \sum_{k=i+1}^{j-1} \alpha_{ijk}\text{\sc combine}\left(\hinside{ik} \oplus \hinside{kj}\right) \text{ where } \bm\alpha_{ij} = \text{\attendfn{}}\left(\begin{bmatrix*}\hinside{i(i+1)} \oplus \hinside{(i+1)j}\\\ldots\\\hinside{i(j-1)} \oplus \hinside{(j-1)j}\end{bmatrix*}\right)\]

\noindent where $\oplus$ denotes vector concatenation and $\bm\alpha_{ij}$ is a vector of non-negative weights such that $\sum_{k=i+1}^{j-1} \alpha_{ijk} = 1$.\footnote{\attendfn{} is implemented with \textit{additive attention} \citep{bahdanau_neural_2015}, a standard deep-learning method of modeling weighted judgments with parameterized functions, $\text{\attendfn{}}(\Xb) = \sigma(\vb^\top \text{tanh}(\Xb\Wb))$, where $\sigma$ is the softmax function, $\sigma(\xb) = \left[\frac{e^{x_1}}{\Sigma_j e^{x_j}}, \frac{e^{x_2}}{\Sigma_j e^{x_j}}, ...\right]$.} These weights can be thought of as probabilities, serving to indicate which subexpressions $w_{i:k}$ and $w_{k:j}$ likely form the expression $w_{i:j}$. 

The outside vectors are analogously defined:

\vspace{-5mm}

\[\houtside{ij}  = \sum\limits_{m=j+1}^{|S|}\alpha_{imj}\text{\text{\sc split}}^R\left(\houtside{im} \oplus \hinside{jm}\right) + 
                            \sum\limits_{m=0}^{i-1}\alpha_{mji}\text{\text{\sc split}}^L\left(\houtside{mj} \oplus \hinside{mi}\right)\\\]

\noindent On analogy with $N_{ij} \cap O_{ij}$ providing a complete picture of $w_{i:j}$ in the context of $S$, we then take the concatenation $\hinsideoutside{ij} = \hinside{ij} \oplus \houtside{ij}$ as the representation for $w_{i:j}$, using it to predict arbitrary labels on $w_{i:j}$. In \S\ref{sec:experiments}, we describe how we train $w_{0:|S|}$ to predict the acceptability of $S$. But our aim is to go beyond merely predicting acceptability: we furthermore aim to induce a coherent symbolic typing on all nodes.

\section{Integrating symbolic types}
\label{sec:computational-model}

Building on the model described in \S\ref{sec:background}, we propose a framework for jointly inferring a sentence's syntactic structure and a coherent mapping from that syntactic structure to semantic types from arbitrary behavioral data. To the vector space syntactic parser described in \S\ref{sec:background}, we add two components: (i) a vector space \textit{interpretation function} that implements $\denotation{\cdot}$ as a parameterized function \textsc{interpret} from the syntactic parser space $\parserspace$ to a denotation space $\interpretationspace = \reals^{M_\text{interp}}$; and (ii) a vector space \textit{type grammar} that takes vector representations produced by the interpretation function and returns the associated symbolic type (or a distribution thereon). 

The high-level idea behind our framework is to train the parser, the interpretation function, and the type grammar to produce an assignment of types to expressions that is coherent in the sense that some type (possibly, but not necessarily, a specific type, like $\st$) can be derived for the entire sentence from the types assigned to its subexpressions. We first describe the type grammar (\S\ref{ssec:type-grammar}) and then describe how we enforce type coherence during training (\S\ref{ssec:interpreter-type-regularizer}). Throughout, we assume a standard recursive definition for the set of types $\types$, given primitive types $\primitivetypes$---e.g. $\entity$, $\state$, and $\truthvalue$.

\vspace{-5mm}

\[\types_0 \equiv \primitivetypes \hspace{4em} \types_i \equiv \left[\bigcup_{j = 0}^{i-1} \types_j\right] \times \directions \times \left[\bigcup_{j = 0}^{i-1} \types_j\right] \hspace{4em} \types \equiv \bigcup_{i=0}^\infty \types_i\]

\noindent where the set of type constructors $\directions$ is a singleton here, but could be straightforwardly extended---e.g. for representing directed types, as in standard and modal syntactic CCG \citep{steedman_syntactic_2000, baldridge_lexically_2002}, or more exotic types, as in generative lexicon theory \citep{asher_lexical_2011,pustejovsky_type_2013,asher_type_2013}.

Throughout this section, we focus only on structural aspects of our framework, specifying how its components hang together. We defer discussion of how we train the parameters of each component to carry out its intended function to \S\ref{sec:experiments}.

\subsection{Type grammar}
\label{ssec:type-grammar}

The type grammar consists of four components: (i)  a set of \textit{primitive type embeddings} $\xb_t$ for primitive types $t \in \primitivetypes$ and \textit{type constructor embeddings} $\xb_{d}$ for $d \in \directions$; (ii) a \textit{type encoder} (\S\ref{sssec:type-encoder}), which maps symbolic types $t \in \types$ to their vector space \textit{type embeddings} $\bm\tau_t$; (iii) \textit{type decoders} (\S\ref{sssec:type-decoder}), which implement vector space combinators and map from the type embeddings corresponding to the combinator arguments to (a distribution over) the resulting types; and (iv) a \textit{combinatory controller} (\S\ref{sssec:combinatory-controller}), which determines which combinator to use for combining a pair of types.\footnote{Our actual implementation of the type encoder uses a stacked binary tree long-short term memory network (binary tree bi-LSTMs; \citealt{le_compositional_2015,tai_improved_2015,miwa_end--end_2016}) and the implementation of the type decoders uses a unidirectional variant. We give a somewhat simplified treatment below that ignores bidirectionality, stacking, and the distinction between hidden states and cell states because they are not relevant to understanding the core ideas.}

\subsubsection{Type encoder}
\label{sssec:type-encoder}

The type encoder views types as binary trees---e.g. $\eet$ and $\montaguetype{\montaguetype{\entity}{\entity}}{\truthvalue}$ are viewed as right-branching and left-branching trees, respectively, both with leaves $\entity$, $\entity$, and $\truthvalue$. It recursively maps these binary type trees to type embeddings by defining the embedding $\bm\tau_t$ of complex type $t = [_d\;t_0\;t_1]$ in terms of embeddings of its (possibly complex) constituent types $t_0$ and $t_1$ and the type constructor $d$ using parameterized functions \textsc{wrap} and \textsc{construct}.

\vspace{-3mm}

\[
\bm\tau_t = \begin{cases} \text{\sc wrap}\left(\mathbf{x}_{t}\right) & \text{if } t \in \primitivetypes \\
\text{\sc construct}\left(\mathbf{x}_d, \bm\tau_{t_0} \oplus \bm\tau_{t_1}\right) & \text{if } t = [_d\;t_0\;t_1]
\end{cases}
\]

\subsubsection{Type decoders}
\label{sssec:type-decoder}

Type decoders implement combinators of $M$ arguments (here, $M$ is always 1 or 2) by taking $M$ type embeddings and outputting a probability distribution over output types. For instance, a type decoder implementing the identity combinator should take $\bm\tau_t$ and yield a probability distribution that assigns a probability to $t$ of approximately 1, while a type decoder implementing the application combinator should take $\bm\tau_{\montaguetype{t_0}{t_1}}$ and $\bm\tau_{t_0}$ and yield a probability distribution that assigns a probability to $t_1$ of approximately 1. The reason that we need these combinators to produce probability distributions rather than types, is that multiple types may be compatible with a certain expression---e.g. because it is ambiguous---and so, when we apply these decoders to the vector space interpetation of an expression, we need to allow for multiple types---even if, when applied to a type embedding, we only want the decoder to yield the single correct type. 

Type decoders can be thought of as the reverse of encoders, ``reading'' a distribution over symbolic types off of a vector space embedding. Each decoder consists of three parameterized functions: (i) \textsc{structure}, which determines the probability that the decoder produces a primitive type v. a complex type; (ii) \primitivefn{}, which determines which primitive type to select if the decoder chooses to produce a primitive type; and (iii) \textsc{factor}, which determines how to update the decoder's state based on its previous state (by ``factoring out'' the previous decision to recurse from the state).\footnote{When $\directions$ is a non-singleton, a fourth parameterized function is required for selecting the constructor.} The start state of the decoder is always the input type embedding(s). 

For ease of exposition, we describe how a single
type would be sampled from the distribution defined by \textsc{structure}, \primitivefn{}, \textsc{factor}, and the input type embedding(s), though it is also possible to determine the probability of particular types as well as the most likely types using related procedures. Starting with the input embedding(s) $\bm\tau$, the \textsc{sample}($\bm\tau$) procedure chooses to generate a complex type with probability $\theta_\text{complex} = \text{\sc structure}(\bm\tau)$ and a primitive type with probability $1-\theta_\text{complex}$. If it chooses to generate a primitive type, it chooses that primitive type based on primitive type probabilities $\bm\theta_\text{primitive} = \text{\primitivefn{}}(\bm\tau)$ and returns it; otherwise, it returns $\langle \text{\sc sample}(\bm\tau'_0), \text{\sc sample}(\bm\tau'_1)\rangle$, where $\langle\bm\tau'_0, \bm\tau'_1\rangle = \text{\sc factor}(\bm\tau)$.

\subsubsection{Combinatory controller}
\label{sssec:combinatory-controller}

When types are represented symbolically, it is straightforward to determine whether they are in the domain of a combinator and, if not, whether type raising one of the types---i.e. mapping $t \in \types$ to $\montaguetype{\montaguetype{t}{t'}}{t'}$ for some $t' \in \types$---would make them so. When types are represented as vectors, this is not the case (except in the case of the identity combinator, which can apply to any type embedding). 

To deal with this issue, we need some way of selecting (i) which combinator to use for decoding the input; (ii) whether to type-raise one of the children; and (iii) if a particular child is to be raised, which type to raise that child to. In addition, we need to define how raising is carried out in a vector space.

These decisions are made based on a set of \textit{combinatory action types} $\actions$, which are triples specifying which combinator to use and whether to raise the left type, right type or neither, as well as two parameterized functions: (i) \actionfn{}, which selects an action type from $\actions$ for $\bm\tau$ and $\bm\tau'$ based on $\bm\phi^\text{(action)} = \text{\actionfn{}}\left(\bm\tau, \bm\tau'\right)$; and (ii) \raisefn{}, which selects a type with which to raise $\bm\tau$ with probability $\bm\phi^{(raise)} = \text{\raisefn{}}(\bm\tau)$ (e.g. if $\truthvalue$ is selected to raise $\bm\tau_\text{\entity}$, $\bm\tau_{\montaguetype{\montaguetype{\entity}{\truthvalue}}{\truthvalue}}$ should result).\footnote{We only consider raising with primitive types. There are multiple way one might extend vector space type raising to complex types. We take this issue to be somewhat unimportant: it is unclear whether there is a real need for vector space type raising, since the same result might already be achievable by the model by positing ambiguity.} To implement type raising of $\bm\tau_t$ with $t'$, we run the type encoder on a \textit{type embedding tree} $[_{\xb_{d}} [_{\xb_{d}} \bm\tau\; \bm\tau_{t'} ]\;\bm\tau_{t'} ]$ corresponding to the raised type, as though $\bm\tau_t$ were a primitive type embedding.

\subsection{Optimizing type coherence}
\label{ssec:interpreter-type-regularizer}

In order to learn an interpretation function from the syntactic parser space to the denotation space that produces coherent types, we minimize what we term a \textit{type coherence loss} between the type assigned to the interpretation of an expression by the identity combinator with the type produced by combining its children using the combinatory action selected by the combinatory controller. To compute the type coherence loss for the vector space interpretation $\bm\lambda_{ij} = \text{\sc interpret}(\hinsideoutside{ij})$ of expression $w_{i:j}$ and the vector space interpretations $\bm\lambda_{ik} = \text{\sc interpret}(\hinsideoutside{ik})$ and $\bm\lambda_{kj} = \text{\sc interpret}(\hinsideoutside{kj})$ (with $i < k < j$) for a pair of subexpressions $\langle w_{i:k}, w_{k:j}\rangle$ of $w_{i:j}$, we compute the probability distribution over types $\prob^\text{(parent)}$ produced by applying the identity decoder to $\bm\lambda_{ij}$ and the probability distributions over types $\prob^\text{(children)}_a$ produced by conducting combinatory action $a \in \actions$ on $\bm\lambda_{ik}$ and $\bm\lambda_{kj}$. The agreement between $\prob^\text{(parent)}$ and $\prob^\text{(children)}_a$ is computed using cross-entropy $H$.\footnote{We approximate this value by computing the probability distribution over type trees of at most a certain depth (here, depth 4). Alternative methods for comparing probability distributions---e.g. Kullback-Leibler divergence---could be used in place of cross-entropy.} The contribution to this agreement score from action $a$ is then weighted by the action probabilities $\bm\phi^\text{(action)}_{ijk} = \text{\actionfn{}}(\bm\lambda_{ik}, \bm\lambda_{kj})$ assigned by the combinatory controller.

\vspace{-5mm}

\begin{align*}
    \loss^\text{(pair)}_\text{type}(i, j, k) &= \sum_{a \in \actions} \phi^\text{(action)}_{ijka} H\left(\prob^\text{(parent)}(\cdot\;|\;\bm\lambda_{ij}), \prob^\text{(children)}_a(\cdot\;|\;\bm\lambda_{ik}, \bm\lambda_{kj})\right)\\
                                                               &= -\sum_{a\in\actions} \phi^\text{(action)}_{ijka} \sum_{t \in \types} \prob^\text{(parent)}(t\;|\;\bm\lambda_{ij}) \log \prob^\text{(children)}_a(t\;|\;\bm\lambda_{ik}, \bm\lambda_{kj})\\
\end{align*}

\vspace{-8mm}

\noindent This value is small when $\prob^\text{(children)}_a$ assigns probabilities to types that are similar to those assigned by $\prob^\text{(parent)}$. The type coherence for an expression $w_{i:j}$ is then computed by summing over all possible pairs of subexpressions $\langle w_{i:k}, w_{k:j}\rangle$, weighting by the likelihood $\alpha_{ijk}$ that the syntactic parser assigns to that particular pair: $\loss^\text{(expr)}_\text{type}(i, j) = \sum_{k=i+1}^{j-1} \alpha_{ijk} \loss_\text{type}(i, j, k)$. This value is small when the type coherence loss for high probability pairs is small. Finally, the type coherence loss for an entire sentence is computed by summing over the type coherence loss for all expressions: $\loss^\text{(sent)}_\text{type} = \sum_{i=0}^{|S|-2} \sum_{j=i+2}^{|S|} \loss_\text{type}(i, j)$. This value is small when the type coherence loss for all expressions is small, and thus, our aim is to find parameterizations for the syntactic parser, type grammar, and interpretation function that drive $\loss^\text{(sent)}_\text{type}$ down.

\section{Deploying the framework}
\label{sec:experiments}

As a proof-of-concept experiment using our framework, we fit a model that simultaneously aims to predict sentence acceptability and find a coherent typing for the expressions contained in those sentences. Our framework allows us to specify the that grammar has an arbitrary number of primitive types $\primitivetypes$ and type constructors $\directions$ as well as arbitrary combinators $\combinators$. For these experiments, we follow standard treatments building on \citealt{montague_proper_1973} in assuming three primitive types ($\entity$, $\state$, and $\truthvalue$) and a single type constructor. For the set of combinators $\combinators$, we assume (undirected) application---where \textsc{apply}$(t_0, \montaguetype{t_0}{t_1})$ = \textsc{apply}$(\montaguetype{t_0}{t_1}, t_0)$ = $t_1$---and (first-order undirected) composition---where \textsc{compose}$(\montaguetype{t_0}{t_1}, \montaguetype{t_1}{t_2})$ = \textsc{compose}$(\montaguetype{t_1}{t_2}, \montaguetype{t_0}{t_1})$ = $\montaguetype{t_0}{t_2}$. We first describe the acceptability dataset that we train the model on (\S\ref{ssec:data}) and then discuss how we train the model to simultaneously predict acceptability and coherent types (\S\ref{ssec:training}).

\subsection{The MegaAcceptability dataset}
\label{ssec:data}

We train our model to predict the acceptability judgments in the MegaAcceptability dataset \citep{white_computational_2016,white_frequency_accepted}. This dataset contains acceptability judgments for 1,000 clause-embedding verbs---including a variety of cognitive verbs (e.g. \textit{believe}, \textit{forget}, \textit{doubt}), communicative verbs (e.g. \textit{say}, \textit{claim}, \textit{explain}), emotive verbs (e.g. \textit{upset}, \textit{disgust}, \textit{anger}), among many other classes---in 50 different syntactic frames. These frames include simple intransitives (\textsc{np} \underline{\phantom{ver}}), transitives (\textsc{np} \underline{\phantom{ver}} \textsc{np}), and ditransitives (\textsc{np} \underline{\phantom{ver}} \textsc{np} \textsc{np}) as well as a large variety of frames including subordinate clauses either with or without a direct object or prepositional phrase (always headed by \textit{to}). These subordinate clauses vary in terms of their complementizer ($\emptyset$, \textit{that}, \textit{for}, \textit{whether}, and constituent question), the presence of an embedded subject, and embedded tense (past, future, tenseless, \textit{to}-infinitival, bare infinitival, and present participial). (See \citeauthor{white_frequency_accepted} for a full list of frames.)

A sentence is constructed for each verb-frame pair by instantiating all lexical category words with bleached terms to minimize the effects of lexical idiosyncrasies in the acceptability judgments. All noun phrases are instantiated as either \textit{someone} or \textit{something}, all untensed verb phrases as either \textit{do something} or \textit{have something} and tensed verb phrases as \textit{happened}, and sentences as \textit{something (would) happen(ed)}.
\ex. \a. \label{ex:believe-frame} \textit{believe} + \textsc{np} \underline{\phantom{ver}} \textit{that} \textsc{s}
            $\rightarrow$ Someone believed that something happened.            
     \b. \label{ex:wonder-frame} \textit{ask} + \textsc{np} \underline{\phantom{ver}} \textit{whether} \textsc{s}
            $\rightarrow$ Someone asked whether something happened.
     \b. \label{ex:ask-np-to-frame} \textit{force} + \textsc{np} \underline{\phantom{ver}} \textsc{np} to \textsc{vp}
            $\rightarrow$ Someone forced someone to do something.
     \b. \label{ex:tell-np-which-to-frame} \textit{tell} + \textsc{np} \underline{\phantom{ver}} \textsc{np} \textit{which thing} to \textsc{vp}
            $\rightarrow$ Someone told sm. which thing to do.

Each sentence is rated by 5 participants on a 1-7 scale. We use the normalized variant of these judgments provided by \cite{white_frequency_accepted}, which adds seven verbs not in the original data and maps each sentence to a single real-valued acceptability value in a way that accounts for annotator bias and annotator quality.

\subsection{Training the model}
\label{ssec:training}

Our model is trained using the Adam optimizer in three stages: (i) syntactic parser training, (ii) type grammar training, and (iii) interpreter training.

\subsubsection{Syntactic parser training}

The parameters of the syntactic parser---i.e. the parameters of \textsc{unary}, \textsc{combine}, \textsc{split}$^R$, and \textsc{split}$^L$ discussed in \S\ref{sec:background}---are trained to predict the normalized acceptability judgments found in MegaAcceptability, minimizing the mean-squared error (as in standard linear regression). The RoBERTa system \citep{liu_roberta_2019} is used to map $w_0...w_{|S|-1}$ to input vectors $\xb_0...\xb_{|S|-1}$. The acceptability prediction is done by passing the vector representation $\hinsideoutside{0|S|}$ for the entire sentence into a parameterized function \textsc{acceptability}, whose parameters are jointly trained with those of the parser.

\subsubsection{Type grammar training}

The parameters of the type grammar---i.e. the parameters of the encoder's \textsc{wrap} and \textsc{construct}, the decoders' \textsc{structure}, \textsc{primitive}, and \textsc{factor}, and the combinatory controller's \textsc{action} and \textsc{raise}---are trained in three stages. 

\textbf{Encoder and identity decoder.} First, the type encoder and identity combinator are trained as an \textit{autoencoder}---i.e. so that the probability distribution over types produced by encoding $t$ as $\bm\tau_t$ then applying the identity combinator to $\bm\tau_t$ assigns high probability to $t$. This training is carried out on randomly sampled types. 

\textbf{Application and composition decoders.} The remaining combinators in the grammar are similarly trained. For each combinator, pairs of types are randomly sampled and automatically processed so that they are viable inputs to the corresponding combinator---e.g. ensuring that only pairs of the form $\montaguetype{t_0}{t_1}$ and $t_0$ (or the reverse) are input to application and only pairs of the form $\montaguetype{t_0}{t_1}$ and $\montaguetype{t_1}{t_2}$ (or the reverse) are input to composition. To train each combinator, the correct outputs for each of these pairs is first determined symbolically based on the combinator---e.g. for application, the output for the pair $\montaguetype{t_0}{t_1}$ and $t_0$ (or the reverse) is $t_1$---then each element of the pair is encoded using the type encoder. The parameters of the combinator are then trained to produce a high probability for the correct output when applied to the concatenation of the encoded pair. 

\textbf{Combinatory controller.} The combinatory controller is trained with a similar sampling procedure to the decoder training. The sampling procedure is extended so that when processing the type pairs into viable inputs, we include the possibility of type-raising one of the inputs with respect to a primitive type before combinator application. Viable combinator action types are symbolically determined for each input pair---e.g. $\montaguetype{t_0}{t_1}$ and $t_0$ could be combined with application and no type raising or with application and type raising $t_0$ to $\montaguetype{\montaguetype{t_0}{t_1}}{t_1}$. The parameters of the combinatory controller are then trained to produce a high probability for viable combinatory action types when applied to the concatenation of the input pair.

\subsubsection{Interpreter training}

The parameters of \textsc{interpret} are trained to minimize the type coherence loss $\loss^\text{(sent)}_\text{type}$ (\S\ref{ssec:interpreter-type-regularizer}) and a \textit{type constraint loss}. The type constraint loss
encodes standard assumptions about the types that particular lexical items map to. Here, we enforce that sentence denotations have proposition types $\st$, quantificational noun phrase denotations (\textit{someone}, \textit{something}) have quantifier types $\montaguetype{\est}{\st}$, and verb phrases \textit{do something}, \textit{have something}, \textit{happen}, and \textit{happened} denote properties of individuals $\est$.\footnote{Without these constraints, we found that the model would learn a trivial grammar---e.g. every constituent denotes $\montaguetype{\truthvalue}{\truthvalue}$ and the combinatory controller assigns composition a high probability. Crucially, we do not enforce that particular clausal complements decode to particular types, since we aim to induce those types.} To enforce these constraints, we train the interpretation function so that applying the identity combinator to the denotation embedding vector $\bm\lambda_{ij}$ of a span matching one of these expressions assigns high probability to the corresponding type. This training procedure is only applied to the subset of the data where the normalized acceptability value is in the 90th percentile or greater, so that we do not attempt to derive $\st$ for sentences that should not have such a type. This threshold is fairly stringent---the least acceptable sentence above the threshold is \ref{ex:least}---and likely removes acceptable sentences. 

\ex. Someone distrusted someone that something happened. \label{ex:least}

Removing acceptable sentences is fine for our purposes, since it is more important to avoid forcing the model to produce coherent types for unacceptable sentences.

\subsection{Predicting types}

To find the most likely type for each subexpression of a sentence, we use a probabilistic parsing algorithm based on the standard supertag-factored A* CCG parsing algorithm~\citep{lewis_ccg_2014,lewis_lstm_2016}. The specific details of this algorithm are not relevant for current purposes.

\section{Results}
\label{sec:results}

We will first take a look at how well the parser does in predicting acceptability and explore some of the syntactic representations it learns. This is mainly to show that the parser is learning something that could reasonably be thought of as a syntactic representation. Then, we  will dive into the types that are learned by our model, drawing two contrasts: (i) what the model infers for both declarative and interrogative complement clauses and (ii) what it infers for finite and infinitival complements.

\subsection{Acceptability}
\label{ssec:acceptability}

We use $k$-fold cross-validation to evaluate our model's ability to predict the normalized acceptability of sentences it has not seen. In cross-validation~\citep[see][\S7]{hastie_elements_2009}, the data is partitioned into $k$ parts (here, $k$ = 5), and for each of the cells, a model is trained on all but that cell and then its performance is evaluated on that cell. Averaging across five validation folds, our model obtains a Pearson correlation of $\rho$ = 0.71 (95\% CI=[0.69, 0.73]). This correlation is extremely close to the one reported by \citet{white_frequency_accepted}: trained linguists agree on a subset of the MegaAttitude sentences with a Spearman correlation of 0.70 (95\% CI=[0.62,0.78]). This result suggests that the models agrees with the normalized scores at about the same level that trained linguists agree with each other.

\subsection{Syntactic Representations}
\label{ssec:syntax-plots}

\begin{figure}[t]
    \centering
    \includegraphics[width=\textwidth]{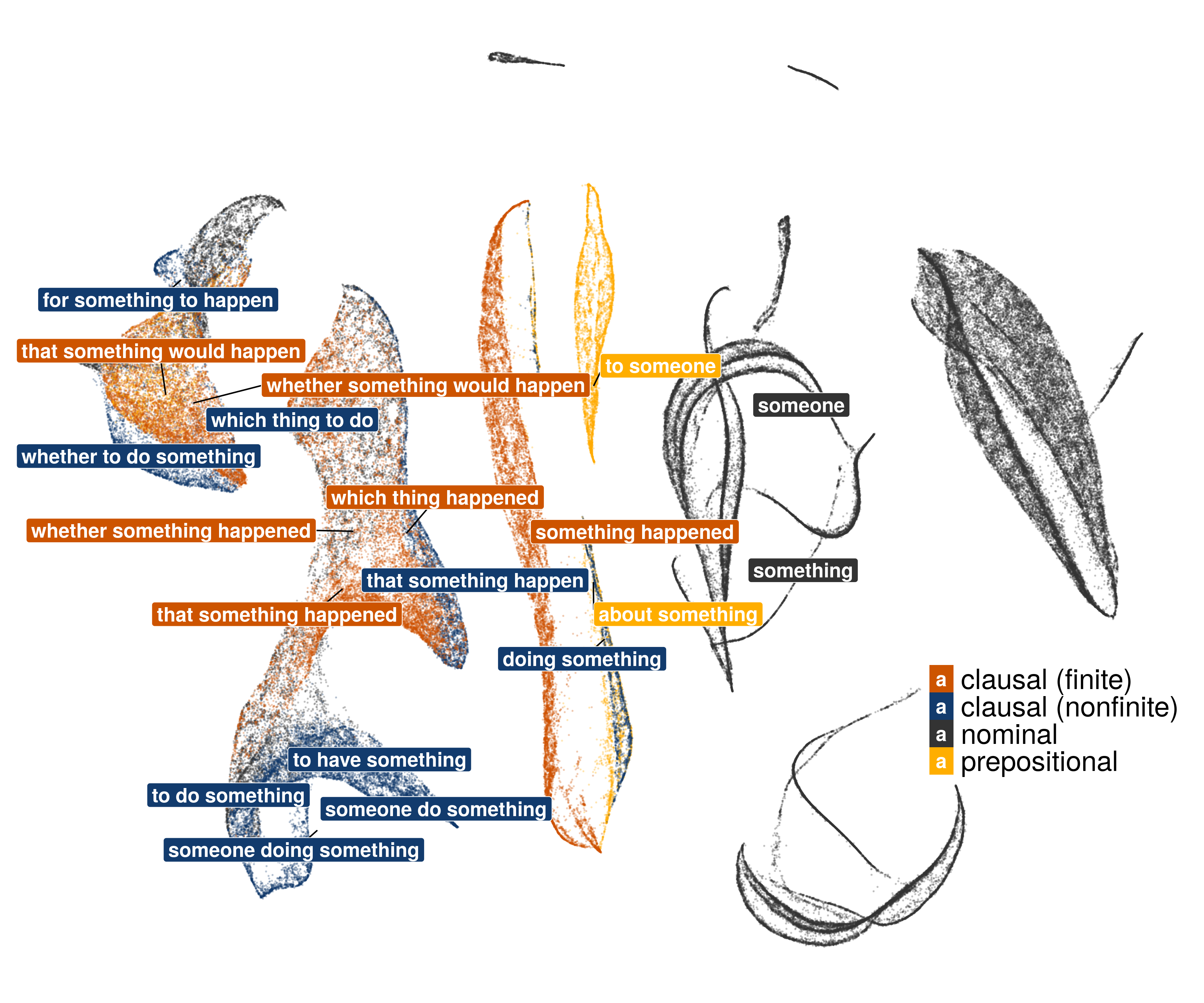}
    \vspace{-14mm}
    \caption{The UMAP visualization for the parser's representation for different expressions colored by coarse-grained expression types. Labels show the mean location of the particular expression.
    } \label{fig:span-embedding-space} 
\end{figure}

Next, we investigate the vector representations for particular constituents. To do this, we run the syntactic parser on all sentences in MegaAcceptability and extract the vector representations for various nominal, prepositional, and  clausal expressions in all contexts they appear. This yields as many vectors for an expression as it has unique tokenings across the dataset---e.g. \textit{whether something happened} appears once in each of 6 frames and so, across 1,007 verbs, we obtain 6,042. It is important to extract a vector for every context in which a particular constituent appears because the outside embedding $\houtside{ij}$ differs for the same expression $w_{i:j}$ depending on $w_{0:i}, w_{j:|S|}$.

To visualize how these vectors are arranged, we apply the UMAP dimensionality reduction technique~\citep{mcinnes_umap_2018} to map them to two dimensions in a way that preserves the relationships between vectors in the high dimensional space. Figure~\ref{fig:span-embedding-space} plots the results. We see a broad split between clausal expressions on the left and nominal and prepositional expressions on the right. The cluster in the upper left contains finite clauses with a future modal as well as infinitival interrogatives and \textit{for}-\textit{to} clauses. The absence of other infinitivals from this cluster---e.g. subjectless infinitivals and bare infinitivals with and without subjects, which are tightly grouped in a separate cluster---may suggest that the model is formally representing some combination of having a overt subject (or perhaps a complementizer) and containing a modal---in this case, \textit{would} or \textit{to} \citep[][cf. \citealt{stowell_tense_1982,ogihara_tense_1996,martin_null_2001,katz_temporal_2001,pearson_semantics_2016}; and see also \citealt{grano_control_2012,grano_control_2017,williamson_temporal_2019}]{bhatt_covert_1999,wurmbrand_tense_2014}. 

The cluster next to that contains finite clauses---both declarative and interrogative---and tenseless \textit{that} clauses in the upper portion and subjectless and bare infinitivals in the lower portion. The fact that the tenseless \textit{that} clause clusters with the finites may suggest that this grouping is mainly determined by the presence of an overt complementizer. This is bolstered by the fact that the finite clause without a complementizer is contained within its own long thin region just to the right.

\subsection{Types}
\label{ssec:decoded-types}

\begin{figure}[t]
    \centering
    \includegraphics[width=\textwidth]{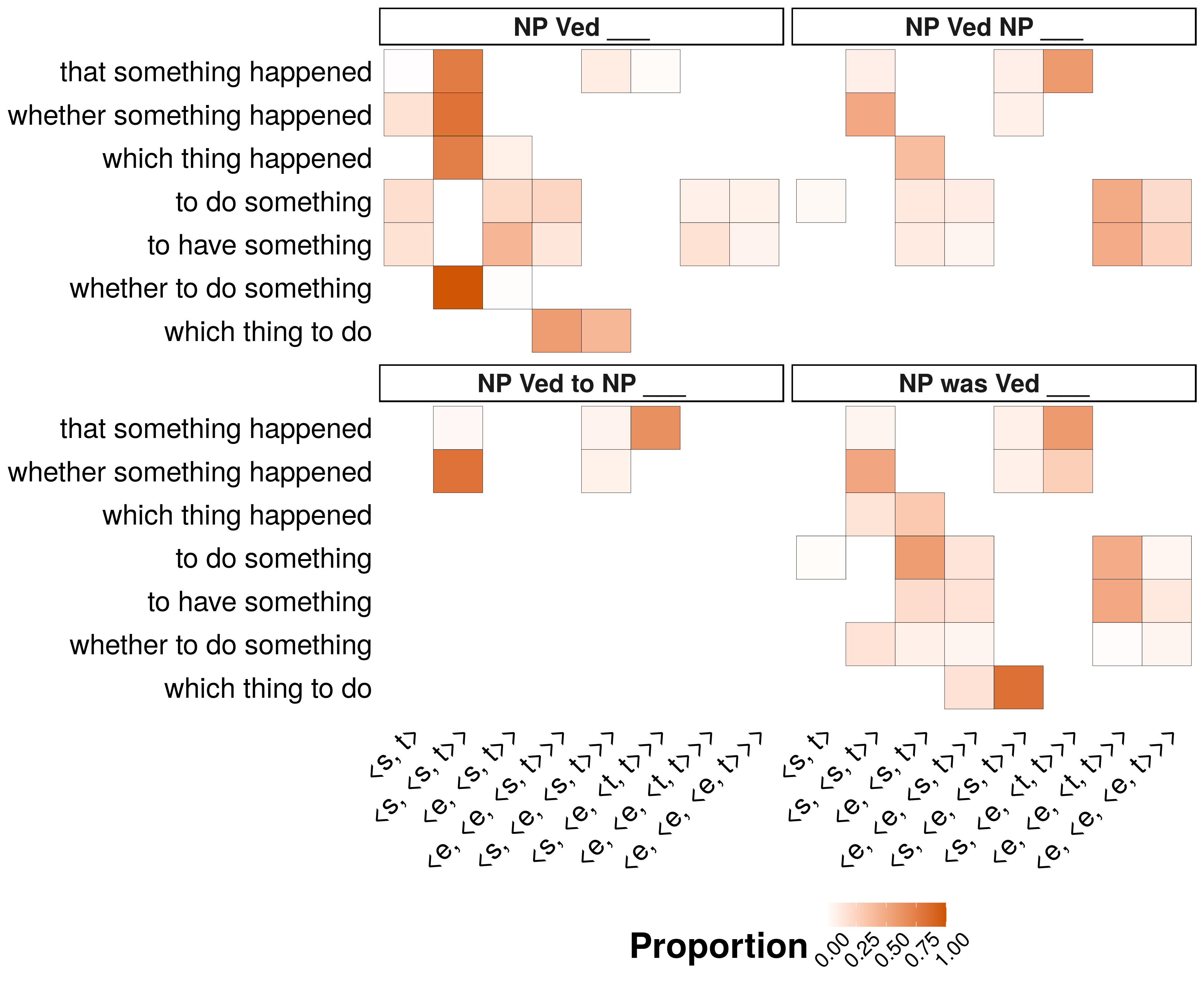}
    \caption{\label{fig:complement-types} The types most frequently decoded for each expression across verbs.}
    \vspace{-12mm}
\end{figure}

Having established that the syntactic representations our parser learns are reasonable, we now turn to the semantic types our model assigns to different expressions within the high acceptability set used to train the type coherence loss (90th percentile acceptability or above). To obtain these types, we apply the decoding algorithm (see \S\ref{sec:experiments}) to the vector for an expression in each context in which it is found---i.e. the same vectors analyzed in \ref{ssec:syntax-plots}---in order to obtain the most likely type(s).\footnote{An alternative, more computationally intensive method is to decode the type for the expression in the context of the entire sentence. We leave this for future work.} 

For reasons of space, we focus in particular on the types assigned to expressions along two axes of variation: interrogative v. declarative and finite v. infinitival. Figure~\ref{fig:complement-types} shows the proportion of tokenings of a particular expression (\textit{y}-axis) that are assigned particular types (\textit{x}-axis) in a particular context (facets) across verbs.\footnote{Indicative of how stringent the 90th percentile threshold is, some rows for some contexts are blank because the relevant expression does not show up in that context within the set of items we investigate.}

In intransitive contexts (\textsc{np} \textsc{v}\textit{ed} \underline{\phantom{ver}}), declaratives (\textit{that something happened}), finite interrogatives (\textit{whether something happened}, \textit{which thing happened}), and infinitival polar interrogatives (\textit{whether to do something}) all heavily favor the $\sst$ type, suggesting that our model infers that they denote questions represented as the set of their complete answers (see \citealt{hamblin_questions_1958,groenendijk_studies_1984} among others; cf. \citealt{uegaki_interpreting_2015}). Less frequently, declaratives and finite polar interrogatives are assigned the proposition type $\st$, and constituent interrogatives are assigned the $\est$ type, suggesting a functional question type \citep{krifka_questions_2011} where the entity argument denotes the ``missing piece'' corresponding to the \textsc{wh}-phrase~\citep{hintikka_semantics_1976,berman_semantics_1991}. Interestingly, polar interrogative that are assigned $\sst$ tend to be complements of responsive predicates---i.e. predicates that take both declaratives and interrogatives, like \textit{know}, \textit{discover}, \textit{find out}, and \textit{realize}---while polar interrogatives that are assigned the proposition type $\st$ tend to be rogative predicates \citep{lahiri_questions_2002}---i.e. predicates that only take interrogatives (or are at least marginal with declaratives), like \textit{wonder} and \textit{ask}. There are potentially interesting exceptions to this generalization, though: \textit{doubt} is responsive, but its polar question complement is assigned a proposition type---possibly related to the intuitive equivalence between \textit{doubt whether} and \textit{doubt that} \citep{karttunen_doubt_1977}.

In contrast to finite constituent interrogatives, infinitival constituent interrogatives (\textit{which thing to do}) tend to favor the type corresponding to properties of relations between individuals $\eest$, with the type corresponding to intensionalized properties of individuals $\sest$ a close second. At least, one of the entity types in both might be understood as the missing subject, and in the case of $\eest$, the second might correspond to the \textsc{wh}-phrase, as in a functional question. It may be that this distinction is a real one---between representing constituent interrogatives as functional questions with a missing argument or as partition questions with a missing argument---but in analyzing the distribution over types for particular verbs' complements, we found that there are very few examples where our model assigns $\sest$ for its highest probability prediction but not $\eest$ as one of the runner-up predictions and vice-versa. This pattern may indicate uncertainty in the decoder as to which is the correct analysis or it may be that the decoder has trouble distinguishing between the two analyses because they are so close to each other in terms of tree edits---differing by a single primitive type. One potential argument against this possibility is that the decoder appears to be sensitive to the context (and/or the verbs that appear in said context) when deciding between these types: it slightly prefers $\eest$ in intransitive contexts but strongly prefers $\sest$ in passivized transitive contexts (\textsc{np} \textit{was} \textsc{v}\textit{ed} \textsc{np} \underline{\phantom{ver}}).

Finally, the distribution over types for the non-interrogative subjectless infinitivals (\textit{to do something}, \textit{to have something}) shows a more diffuse distribution on types, relatively evenly weighting the proposition type $\st$, the type corresponding to properties of individuals $\est$, and the type corresponding to properties of relations between individuals $\eest$.\footnote{As for infinitival constituent interrogatives, we suspect that the other types $\eett$ and $\eeet$ may be hard for the decoder to distinguish from $\eest$.} The proposition type is expected under theories that posit a covert pronoun in subject position of the infinitival (\citealt{rosenbaum_grammar_1967} \textit{et seq}) and tends to be associated with verbs of appearance (e.g. raising verbs: \textit{appear} and \textit{seem}), verbs of claiming and pretense (e.g. \textit{claim}, \textit{pretend}), verbs of choice (e.g. \textit{decide}, \textit{choose}), and emotives (e.g. \textit{like}, \textit{love}).\footnote{Indeed, many (though not all) of the verbs that are assigned the proposition type are exactly those that would be full tense phrases in \citeauthor{wurmbrand_tense_2014}'s (\citeyear{wurmbrand_tense_2014}) theory of infinitival complements.}  The property-of-individuals type is expected under theories that do not assume such a covert pronoun \citep{bach_control_1979, chierchia_topics_1984,dowty_recent_1985} and tends to be associated with complements of aspectual verbs (e.g. \textit{start}, \textit{stop}), intention verbs (e.g. \textit{try}, \textit{intend}, \textit{plan}), desire verbs (e.g. \textit{want}, \textit{wish}), and some cognitive verbs (e.g. \textit{remember}, \textit{forget}). 

The interpretation of the property-of-relations-between-individuals type is less clear but its assignment appears to correlate with whether the selecting verb can take direct object or PP arguments that can go unexpressed---e.g. \textit{say}, \textit{argue}, \textit{advise}, \textit{look} (for), \textit{long} (for)---and so the model may be learning to ``pack'' those unexpressed arguments into the clause. This possibility is bolstered by the pattern seen in the transitive (\textsc{np} \textsc{v}\textit{ed} \textsc{np} \underline{\phantom{ver}}, \textsc{np} \textit{was} \textsc{v}\textit{ed} \textsc{np} \underline{\phantom{ver}}) and preposition phrase (\textsc{np} \textsc{v}\textit{ed to} \textsc{np} \underline{\phantom{ver}}) contexts, where the possible types for the declarative and finite polar interrogative remain largely the same, though the proportions shift somewhat toward types that have an ``extra'' \entity{}. This additional \entity{} might suggest a content individual \citep{kratzer_decomposing_2006,moulton_natural_2009,moulton_cps:_2015,bogal-allbritten_building_2016}---i.e. an entity whose content is constrained by the content of the clause---though this would be surprising in light of the correlation with direct object- and preposition-taking behavior. More likely, we suspect, is that the model treats clauses as polysemous between relational and non-relational variants, where the relational variant takes the denotation of a direct object or prepositional phrase as an argument and relates it to the content of the clause via some free relation \citep{partee_uniformity_1997} that is further constrained by the verb. This possibility seems perverse until noting that, like the two noun phrase internal arguments in ditransitives, the noun phrase-clause complex can be ``non-constituent'' coordinated.

\ex. 
\a. Jo told Bo that Mo left and Mo that Bo left.
\b. Jo told Bo to leave and Mo to stay.

The main difference between standard CCG analyses \citep[see][]{steedman_syntactic_2000} and what the model appears to be doing is that, unlike standard analyses, where the coordinated constituent is combined via type-raising followed by composition, the model appears to be imbuing the clause with argument-taking behavior. As such, an alternative take is that the model is attempting to approximate something like a neo-\citeauthor{davidson_logical_1967}ian analysis \citep[see][]{parsons_events_1990} without primitive event types or a predicate modification rule \citep[see][]{heim_semantics_1998} for ensuring compositional interpretation. Future work might test this possibility within our framework by adding an additional primitive type for events (with concomitant alteration of the type constraints) as well as an additional type decoder for predicate modification.

\section{Conclusion}
\label{sec:conclusion}

We have presented a computational modeling framework for inducing combinatory categorial grammars from arbitrary behavioral data. This framework provides the analyst fine-grained control over the assumptions that the induced grammar should conform to: (i) what the primitive types are; (ii) how complex types are constructed; (iii) what set of combinators can be used to combine types; and (iv) whether (and to what) the types of some lexical items should be fixed.  As a proof-of-concept experiment, we deployed our framework for use in distributional analysis. We investigated the induced grammar, finding that it assigns a highly interpretable system of types---even for complex clausal expressions of various forms.

This work is only the tip of the iceberg in terms of how our framework might be used in semantic theory-building and evaluation. There are at least three promising areas for future investigation.

\textbf{Jointly training on multiple distinct datasets.} A major benefit of our framework is that it can be used with arbitrary behavioral data. This means not only fitting models to particular datasets, but also using our framework to synthesize multiple datasets into a single induced grammar. In preliminary experiments, we have used our framework to jointly predict acceptability using the syntactic parser representations $\hinsideoutside{0|S|}$ (as in this paper) as well as veridicality judgments from the MegaVeridicality data \citep{white_role_2018,white_lexicosyntactic_2018}---e.g. that \textit{Jo proved that Bo left} entails that \textit{Bo left}, while \textit{Jo suggested that Bo left} does not---using the denotation representations $\bm\lambda_{0|S|}$. We have found that it is possible to jointly  predict veridicality and acceptability judgments at native speaker levels, though the types we induce are less interpretable than those presented here.  

Beyond behavioral data, it is also possible to use our framework to jointly train on corpus data, building on existing deep learning-based grammar induction methods using only the syntactic parser component of our framework \citep{drozdov_unsupervised_labeled_2019,drozdov_unsupervised_latent_2019}. We are currently experimenting with joint syntactic and semantic CCG induction from corpus data by expanding the set of type constructors and combinators from undirected to directed variants and enforcing additional constraints to capture \textit{combinatory type transparency} \citep[see][]{steedman_syntactic_2000}.  

\textbf{Alternative grammatical assumptions.} Another major benefit of our framework is that primitive types, type constructors, combinators, and type constraints are highly tunable: the analyst need merely specify the set of primitive types, set of constructors, type constraints, and combinator behavior they are interested in. As noted in \S\ref{sec:results}, one potentially interesting direction is to investigate the grammars induced under neo-Davidsonian assumptions. A note of caution is warranted here: we have not presented experiments probing the limits of our frameworks capabilities in terms of the sorts of grammars they can reliably induce. Further work is necessary.

\textbf{Inducing logical form.} Finally, our framework opens up the possibility of not only assigning types to expressions, but also potentially full logical forms. As for types, this induction might be set up as a problem of decoding symbolic expressions conforming to the syntax of some logic from the interpretation vectors $\bm\lambda_{ij}$.


\bibliography{montague_grammar_induction}

\begin{thebibliography}{77}
\providecommand{\natexlab}[1]{#1}
\providecommand{\url}[1]{\texttt{#1}}
\providecommand{\urlprefix}{}
\expandafter\ifx\csname urlstyle\endcsname\relax
  \providecommand{\doi}[1]{doi:\discretionary{}{}{}#1}\else
  \providecommand{\doi}{doi:\discretionary{}{}{}\begingroup
  \urlstyle{rm}\Url}\fi

\bibitem[{An \& White(2020)}]{an_lexical_2020}
An, Hannah \& Aaron White. 2020.
\newblock The lexical and grammatical sources of neg-raising inferences.
\newblock \emph{Proceedings of the {Society for Computation in Linguistics}}
  3(1). 220--233.
\newblock \doi{10.7275/yts0-q989}.

\bibitem[{Asher(2011)}]{asher_lexical_2011}
Asher, Nicholas. 2011.
\newblock \emph{Lexical {Meaning} in {Context}: {A} web of words}.
\newblock Cambridge: Cambridge University Press.

\bibitem[{Asher \& Pustejovsky(2013)}]{asher_type_2013}
Asher, Nicholas \& James Pustejovsky. 2013.
\newblock A {Type} {Composition} {Logic} for {Generative} {Lexicon}.
\newblock In James Pustejovsky, Pierrette Bouillon, Hitoshi Isahara, Kyoko
  Kanzaki \& Chungmin Lee (eds.), \emph{Advances in {Generative} {Lexicon}
  {Theory}} Text, {Speech} and {Language} {Technology}, 39--66. Dordrecht:
  Springer Netherlands.
\newblock \doi{10.1007/978-94-007-5189-7_3}.

\bibitem[{Bach(1979)}]{bach_control_1979}
Bach, Emmon. 1979.
\newblock Control in {Montague} {Grammar}.
\newblock \emph{Linguistic Inquiry} 10(4). 515--531.
\newblock \urlprefix\url{https://www.jstor.org/stable/4178132}.

\bibitem[{Bahdanau et~al.(2015)Bahdanau, Cho \& Bengio}]{bahdanau_neural_2015}
Bahdanau, Dzmitry, Kyunghyun Cho \& Yoshua Bengio. 2015.
\newblock Neural machine translation by jointly learning to align and
  translate.
\newblock In Yoshua Bengio \& Yann LeCun (eds.), \emph{3rd {International}
  {Conference} on {Learning} {Representations}, {ICLR} 2015, {Conference}
  {Track} {Proceedings}}, San Diego, CA, USA.
\newblock \urlprefix\url{http://arxiv.org/abs/1409.0473}.

\bibitem[{Baker(1979)}]{baker_trainable_1979}
Baker, James~K. 1979.
\newblock Trainable grammars for speech recognition.
\newblock \emph{The Journal of the Acoustical Society of America} 65(S1).
  S132--S132.
\newblock \doi{10.1121/1.2017061}.

\bibitem[{Baldridge(2002)}]{baldridge_lexically_2002}
Baldridge, Jason. 2002.
\newblock \emph{Lexically specified derivational control in {C}ombinatory
  {C}ategorial {G}rammar}.
\newblock Edinburgh: University of Edinburgh {PhD} {Thesis}.

\bibitem[{Baroni et~al.(2014)Baroni, Bernardi \&
  Zamparelli}]{baroni_frege_2014}
Baroni, Marco, Raffaella Bernardi \& Roberto Zamparelli. 2014.
\newblock Frege in space: {A} program for compositional distributional
  semantics.
\newblock \emph{Linguistic Issues in Language Technology} 9(6). 5--110.

\bibitem[{Berman(1991)}]{berman_semantics_1991}
Berman, Stephen~Robert. 1991.
\newblock \emph{On the semantics and logical form of wh-clauses}.
\newblock Amherst, MA: University of Massachusetts {PhD} {Thesis}.

\bibitem[{Bhatt(1999)}]{bhatt_covert_1999}
Bhatt, Rajesh. 1999.
\newblock \emph{Covert modality in non-finite contexts}.
\newblock Philadelphia, PA: University of Pennsylvania {PhD} {Thesis}.

\bibitem[{Bisk \& Hockenmaier(2012{\natexlab{a}})}]{bisk_induction_2012}
Bisk, Yonatan \& Julia Hockenmaier. 2012{\natexlab{a}}.
\newblock Induction of linguistic structure with {C}ombinatory {C}ategorial
  {G}rammars.
\newblock In \emph{Proceedings of the {NAACL}-{HLT} {Workshop} on the
  {Induction} of {Linguistic} {Structure}}, 90--95. Montr{\'e}al, Canada:
  Association for Computational Linguistics.
\newblock \urlprefix\url{https://www.aclweb.org/anthology/W12-1912}.

\bibitem[{Bisk \& Hockenmaier(2012{\natexlab{b}})}]{bisk_simple_2012}
Bisk, Yonatan \& Julia Hockenmaier. 2012{\natexlab{b}}.
\newblock Simple robust grammar induction with {Combinatory} {Categorial}
  {Grammars}.
\newblock In \emph{Proceedings of the {Twenty}-{Sixth} {AAAI} {Conference} on
  {Artificial} {Intelligence}}, vol.~2 {AAAI}’12, 1643--1649. Toronto,
  Ontario, Canada: AAAI Press.

\bibitem[{Bisk \& Hockenmaier(2013)}]{bisk_hdp_2013}
Bisk, Yonatan \& Julia Hockenmaier. 2013.
\newblock An {HDP} model for inducing {C}ombinatory {C}ategorial {G}rammars.
\newblock \emph{Transactions of the Association for Computational Linguistics}
  1. 75--88.
\newblock \doi{10.1162/tacl_a_00211}.

\bibitem[{Bisk \& Hockenmaier(2015)}]{bisk_probing_2015}
Bisk, Yonatan \& Julia Hockenmaier. 2015.
\newblock Probing the linguistic strengths and limitations of unsupervised
  grammar induction.
\newblock In \emph{Proceedings of the 53rd {Annual} {Meeting} of the
  {Association} for {Computational} {Linguistics} and the 7th {International}
  {Joint} {Conference} on {Natural} {Language} {Processing} ({Volume} 1: {Long}
  {Papers})}, 1395--1404. Beijing, China: Association for Computational
  Linguistics.
\newblock \doi{10.3115/v1/P15-1135}.

\bibitem[{Bogal-Allbritten(2016)}]{bogal-allbritten_building_2016}
Bogal-Allbritten, Elizabeth~A. 2016.
\newblock \emph{Building meaning in {Navajo}}.
\newblock Amherst, MA: University of Massachusetts {PhD} {Thesis}.

\bibitem[{Chierchia(1984)}]{chierchia_topics_1984}
Chierchia, Gennaro. 1984.
\newblock \emph{Topics in the syntax and semantics of infinitives and gerunds}.
\newblock Amherst, MA: University of Massachusetts {PhD} {Thesis}.

\bibitem[{Clark \& Yoshinaka(2016)}]{clark_distributional_2016}
Clark, Alexander \& Ryo Yoshinaka. 2016.
\newblock Distributional learning of context-free and multiple context-free
  grammars.
\newblock In Jeffrey Heinz \& José~M. Sempere (eds.), \emph{Topics in
  grammatical inference}, 143--172. Berlin, Heidelberg: Springer.
\newblock \doi{10.1007/978-3-662-48395-4_6}.

\bibitem[{Davidson(1967)}]{davidson_logical_1967}
Davidson, Donald. 1967.
\newblock The logical form of action sentences.
\newblock In Nicholas Rescher (ed.), \emph{The logic of decision and action},
  81--95. Pittsburgh, PA: University of Pittsburgh Press.
\newblock \doi{10.1093/0199246270.003.0006}.

\bibitem[{Dowty(1985)}]{dowty_recent_1985}
Dowty, David~R. 1985.
\newblock On recent analyses of the semantics of control.
\newblock \emph{Linguistics and Philosophy} 8(3). 291--331.
\newblock \urlprefix\url{https://www.jstor.org/stable/25001209}.

\bibitem[{Drozdov et~al.(2019{\natexlab{a}})Drozdov, Verga, Chen, Iyyer \&
  McCallum}]{drozdov_unsupervised_labeled_2019}
Drozdov, Andrew, Patrick Verga, Yi-Pei Chen, Mohit Iyyer \& Andrew McCallum.
  2019{\natexlab{a}}.
\newblock Unsupervised labeled parsing with deep inside-outside recursive
  autoencoders.
\newblock In \emph{Proceedings of the 2019 {Conference} on {Empirical}
  {Methods} in {Natural} {Language} {Processing} and the 9th {International}
  {Joint} {Conference} on {Natural} {Language} {Processing}
  ({EMNLP}-{IJCNLP})}, 1507--1512. Hong Kong, China: Association for
  Computational Linguistics.
\newblock \doi{10.18653/v1/D19-1161}.

\bibitem[{Drozdov et~al.(2019{\natexlab{b}})Drozdov, Verga, Yadav, Iyyer \&
  McCallum}]{drozdov_unsupervised_latent_2019}
Drozdov, Andrew, Patrick Verga, Mohit Yadav, Mohit Iyyer \& Andrew McCallum.
  2019{\natexlab{b}}.
\newblock Unsupervised latent tree induction with deep inside-outside recursive
  auto-encoders.
\newblock In \emph{Proceedings of the 2019 {Conference} of the {North}
  {American} {Chapter} of the {Association} for {Computational} {Linguistics}:
  {Human} {Language} {Technologies}, {Volume} 1 ({Long} and {Short} {Papers})},
  1129--1141. Minneapolis, Minnesota: Association for Computational
  Linguistics.
\newblock \doi{10.18653/v1/N19-1116}.

\bibitem[{Goldberg(2017)}]{goldberg_neural_2017}
Goldberg, Yoav. 2017.
\newblock \emph{Neural {Network} {Methods} for {Natural} {Language}
  {Processing}}, vol.~10 Synthesis {Lectures} on {Human} {Language}
  {Technologies}.
\newblock Morgan \& Claypool Publishers.
\newblock \doi{10.2200/S00762ED1V01Y201703HLT037}.

\bibitem[{Grano(2012)}]{grano_control_2012}
Grano, Thomas. 2012.
\newblock \emph{Control and restructuring at the syntax-semantics interface}.
\newblock Chicago: University of Chicago {PhD} {Thesis}.

\bibitem[{Grano(2017)}]{grano_control_2017}
Grano, Thomas. 2017.
\newblock Control, temporal orientation, and the cross-linguistic grammar of
  \textit{trying}.
\newblock \emph{Glossa} 2(1). 94.
\newblock \doi{10.5334/gjgl.335}.

\bibitem[{Groenendijk \& Stokhof(1984)}]{groenendijk_studies_1984}
Groenendijk, Jeroen \& Martin Stokhof. 1984.
\newblock \emph{Studies on the semantics of questions and the pragmatics of
  answers}.
\newblock Amsterdam: University of Amsterdam {PhD} {Thesis}.

\bibitem[{Hamblin(1958)}]{hamblin_questions_1958}
Hamblin, Charles~Leonard. 1958.
\newblock Questions.
\newblock \emph{Australasian Journal of Philosophy} 36(3). 159--168.
\newblock \doi{10.1080/00048405885200211}.

\bibitem[{Hastie et~al.(2009)Hastie, Tibshirani \&
  Friedman}]{hastie_elements_2009}
Hastie, Trevor, Robert Tibshirani \& Jerome Friedman. 2009.
\newblock \emph{The {Elements} of {Statistical} {Learning}}.
\newblock New York, NY: Springer-Verlag 2nd edn.
\newblock \doi{10.1007/978-0-387-84858-7}.

\bibitem[{Heim \& Kratzer(1998)}]{heim_semantics_1998}
Heim, Irene \& Angelika Kratzer. 1998.
\newblock \emph{Semantics in {Generative} {Grammar}}.
\newblock Oxford: Blackwell.

\bibitem[{Heinz \& Sempere(2016)}]{heinz_topics_2016}
Heinz, Jeffrey \& José~M Sempere (eds.). 2016.
\newblock \emph{Topics in grammatical inference}.
\newblock Berlin, Heidelberg: Springer.
\newblock \urlprefix\url{10.1007/978-3-662-48395-4}.

\bibitem[{Hintikka(1976)}]{hintikka_semantics_1976}
Hintikka, Jaakko. 1976.
\newblock The semantics of questions and the questions of semantics: Case
  studies in the interrelations of logic, semantics and syntax.
\newblock \emph{Acta Philosophica Fennica} 28(4).

\bibitem[{Karttunen(1977)}]{karttunen_doubt_1977}
Karttunen, Lauri. 1977.
\newblock To doubt whether.
\newblock In \emph{The {CLS} {Book} of {Squibs}}, Chicago Linguistic Society.

\bibitem[{Katz(2001)}]{katz_temporal_2001}
Katz, Graham. 2001.
\newblock ({A})temporal complements.
\newblock In {Caroline Fery} \& {Wolfgang Sternefeld} (eds.), \emph{Audiator
  {Vox} {Sapientiae}}, 240--258. Berlin: Akademie Verlag.

\bibitem[{Kim et~al.(2019)Kim, Rush, Yu, Kuncoro, Dyer \&
  Melis}]{kim_unsupervised_2019}
Kim, Yoon, Alexander Rush, Lei Yu, Adhiguna Kuncoro, Chris Dyer \& G{\'a}bor
  Melis. 2019.
\newblock Unsupervised recurrent neural network grammars.
\newblock In \emph{Proceedings of the 2019 {Conference} of the {North}
  {A}merican {Chapter} of the {Association} for {Computational} {Linguistics}:
  {Human} {Language} {Technologies}, {Volume} 1 ({Long} and {Short} {Papers})},
  1105--1117. Minneapolis, Minnesota: Association for Computational
  Linguistics.
\newblock \doi{10.18653/v1/N19-1114}.

\bibitem[{Kratzer(2006)}]{kratzer_decomposing_2006}
Kratzer, Angelika. 2006.
\newblock Decomposing attitude verbs.
\newblock Talk presented at the workshop in honor of Anita Mittwoch. The Hebrew
  University of Jerusalem.

\bibitem[{Krifka(2011)}]{krifka_questions_2011}
Krifka, Manfred. 2011.
\newblock Questions.
\newblock In Claudia Maienborn, Klaus von Heusinger \& Paul Portner (eds.),
  \emph{Semantics. {An} international handbook of natural language meaning},
  vol.~2, 1742--1758. Berlin: De Gruyter Mouton.

\bibitem[{Kwiatkowksi et~al.(2010)Kwiatkowksi, Zettlemoyer, Goldwater \&
  Steedman}]{kwiatkowksi_inducing_2010}
Kwiatkowksi, Tom, Luke Zettlemoyer, Sharon Goldwater \& Mark Steedman. 2010.
\newblock Inducing probabilistic {CCG} grammars from logical form with
  higher-order unification.
\newblock In \emph{Proceedings of the 2010 {Conference} on {Empirical}
  {Methods} in {Natural} {Language} {Processing}}, 1223--1233. Cambridge, MA:
  Association for Computational Linguistics.
\newblock \urlprefix\url{https://www.aclweb.org/anthology/D10-1119}.

\bibitem[{Kwiatkowski et~al.(2011)Kwiatkowski, Zettlemoyer, Goldwater \&
  Steedman}]{kwiatkowski_lexical_2011}
Kwiatkowski, Tom, Luke Zettlemoyer, Sharon Goldwater \& Mark Steedman. 2011.
\newblock Lexical generalization in {CCG} grammar induction for semantic
  parsing.
\newblock In \emph{Proceedings of the 2011 {Conference} on {Empirical}
  {Methods} in {Natural} {Language} {Processing}}, 1512--1523. Edinburgh,
  Scotland, UK.: Association for Computational Linguistics.
\newblock \urlprefix\url{https://www.aclweb.org/anthology/D11-1140}.

\bibitem[{Lahiri(2002)}]{lahiri_questions_2002}
Lahiri, Utpal. 2002.
\newblock \emph{Questions and {Answers} in {Embedded} {Contexts}}.
\newblock Oxford University Press.

\bibitem[{Le \& Zuidema(2014)}]{le_inside-outside_2014}
Le, Phong \& Willem Zuidema. 2014.
\newblock Inside-outside semantics: A framework for neural models of semantic
  composition.
\newblock In \emph{{NIPS} 2014 {Workshop} on {Deep} {Learning} and
  {Representation} {Learning}}, .

\bibitem[{Le \& Zuidema(2015)}]{le_compositional_2015}
Le, Phong \& Willem Zuidema. 2015.
\newblock Compositional distributional semantics with long short term memory.
\newblock In \emph{Proceedings of the {Fourth} {Joint} {Conference} on
  {Lexical} and {Computational} {Semantics}}, 10--19. Denver, Colorado:
  Association for Computational Linguistics.
\newblock \doi{10.18653/v1/S15-1002}.

\bibitem[{Lewis et~al.(2016)Lewis, Lee \& Zettlemoyer}]{lewis_lstm_2016}
Lewis, Mike, Kenton Lee \& Luke Zettlemoyer. 2016.
\newblock {LSTM} {CCG} parsing.
\newblock In \emph{Proceedings of the 2016 {Conference} of the {North}
  {A}merican {Chapter} of the {Association} for {Computational} {Linguistics}:
  {Human} {Language} {Technologies}}, 221--231. San Diego, California:
  Association for Computational Linguistics.
\newblock \doi{10.18653/v1/N16-1026}.

\bibitem[{Lewis \& Steedman(2014)}]{lewis_ccg_2014}
Lewis, Mike \& Mark Steedman. 2014.
\newblock {A}* {CCG} parsing with a supertag-factored model.
\newblock In \emph{Proceedings of the 2014 {Conference} on {Empirical}
  {Methods} in {Natural} {Language} {Processing} ({EMNLP})}, 990--1000. Doha,
  Qatar: Association for Computational Linguistics.
\newblock \doi{10.3115/v1/D14-1107}.

\bibitem[{Liu et~al.(2019)Liu, Ott, Goyal, Du, Joshi, Chen, Levy, Lewis,
  Zettlemoyer \& Stoyanov}]{liu_roberta_2019}
Liu, Yinhan, Myle Ott, Naman Goyal, Jingfei Du, Mandar Joshi, Danqi Chen, Omer
  Levy, Mike Lewis, Luke Zettlemoyer \& Veselin Stoyanov. 2019.
\newblock {RoBERTa}: A robustly optimized {BERT} pretraining approach.
\newblock \emph{arXiv:1907.11692 [cs]}
  \urlprefix\url{http://arxiv.org/abs/1907.11692}.
\newblock ArXiv: 1907.11692.

\bibitem[{Maas et~al.(2013)Maas, Hannun \& Ng}]{maas2013rectifier}
Maas, Andrew~L, Awni~Y Hannun \& Andrew~Y Ng. 2013.
\newblock Rectifier nonlinearities improve neural network acoustic models.
\newblock In \emph{Proceedings of the 30th {International} {Conference} on
  {Machine} {Learning} ({ICML})}, vol.~30 1, 3.

\bibitem[{Manning \& Schütze(1999)}]{manning_foundations_1999}
Manning, Chris \& Hinrich Schütze. 1999.
\newblock \emph{Foundations of statistical natural language processing}.
\newblock Cambridge, MA: MIT Press.

\bibitem[{Martin(2001)}]{martin_null_2001}
Martin, Roger. 2001.
\newblock Null case and the distribution of {PRO}.
\newblock \emph{Linguistic Inquiry} 32(1). 141--166.

\bibitem[{McInnes et~al.(2018)McInnes, Healy \& Melville}]{mcinnes_umap_2018}
McInnes, Leland, John Healy \& James Melville. 2018.
\newblock {UMAP}: Uniform manifold approximation and projection for dimension
  reduction.
\newblock \urlprefix\url{https://arxiv.org/abs/1802.03426}.

\bibitem[{Miwa \& Bansal(2016)}]{miwa_end--end_2016}
Miwa, Makoto \& Mohit Bansal. 2016.
\newblock End-to-end relation extraction using {LSTMs} on sequences and tree
  structures.
\newblock In \emph{Proceedings of the 54th {Annual} {Meeting} of the
  {Association} for {Computational} {Linguistics}}, 1105--1116. Berlin,
  Germany: Association for Computational Linguistics.

\bibitem[{Montague(1973)}]{montague_proper_1973}
Montague, Richard. 1973.
\newblock The proper treatment of quantification in ordinary {English}.
\newblock In K.~J.~J. Hintikka, J.~M.~E. Moravcsik \& P.~Suppes (eds.),
  \emph{Approaches to natural language}, 221--242. Springer.

\bibitem[{Moon \& White(2020)}]{moon_source_toappear}
Moon, Ellise \& Aaron~Steven White. 2020.
\newblock The source of nonfinite temporal interpretation.
\newblock In \emph{Proceedings of the 50th {Annual} {Meeting} of the {North}
  {East} {Linguistic} {Society}}, 11--24. Amherst, MA: GLSA Publications.

\bibitem[{Moulton(2009)}]{moulton_natural_2009}
Moulton, Keir. 2009.
\newblock \emph{Natural selection and the syntax of clausal complementation}.
\newblock Amherst, MA: University of Massachusetts {PhD} {Thesis}.

\bibitem[{Moulton(2015)}]{moulton_cps:_2015}
Moulton, Keir. 2015.
\newblock {CPs}: {Copies} and compositionality.
\newblock \emph{Linguistic Inquiry} 46(2). 305--342.

\bibitem[{Ogihara(1996)}]{ogihara_tense_1996}
Ogihara, Toshiyuki. 1996.
\newblock \emph{Tense, attitudes, and scope} Studies in {Linguistics} and
  {Philosophy}.
\newblock Springer Netherlands.
\newblock \doi{10.1007/978-94-015-8609-2}.

\bibitem[{Parsons(1990)}]{parsons_events_1990}
Parsons, Terence. 1990.
\newblock \emph{Events in the {Semantics} of {English}: {A} study in subatomic
  semantics}.
\newblock Cambridge, MA: MIT Press.

\bibitem[{Partee(1983/1997)}]{partee_uniformity_1997}
Partee, Barbara~H. 1983/1997.
\newblock Uniformity vs. versatility: The genitive, a case study.
\newblock In Johan van Benthem \& Alice ter Meulen (eds.), \emph{The handbook
  of logic and language}, 464--470. New York: Elsevier.

\bibitem[{Pearson(2016)}]{pearson_semantics_2016}
Pearson, Hazel. 2016.
\newblock The semantics of partial control.
\newblock \emph{Natural Language \& Linguistic Theory} 34(2). 691--738.

\bibitem[{Potts(2019)}]{potts_case_2019}
Potts, Christopher. 2019.
\newblock A case for deep learning in semantics: {Response} to {Pater}.
\newblock \emph{Language} 95(1). e115--e124.

\bibitem[{Pustejovsky(2013)}]{pustejovsky_type_2013}
Pustejovsky, James. 2013.
\newblock Type theory and lexical decomposition.
\newblock In James Pustejovsky, Pierrette Bouillon, Hitoshi Isahara, Kyoko
  Kanzaki \& Chungmin Lee (eds.), \emph{Advances in {Generative} {Lexicon}
  {Theory}} Text, {Speech} and {Language} {Technology}, 9--38. Dordrecht:
  Springer Netherlands.
\newblock \doi{10.1007/978-94-007-5189-7_2}.

\bibitem[{Rosenbaum(1967)}]{rosenbaum_grammar_1967}
Rosenbaum, Peter~S. 1967.
\newblock \emph{The {Grammar} of {English} {Predicate} {Complement}
  {Constructions}}.
\newblock Cambridge, MA: MIT Press.

\bibitem[{Shen et~al.(2018)Shen, Lin, Huang \& Courville}]{shen_neural_2018}
Shen, Yikang, Zhouhan Lin, Chin-wei Huang \& Aaron Courville. 2018.
\newblock Neural {Language} {Modeling} by {Jointly} {Learning} {Syntax} and
  {Lexicon}.
\newblock In \emph{6th {International} {Conference} on {Learning}
  {Representations}, ({ICLR}) {Conference} {Track} {Proceedings}}, Vancouver,
  BC, Canada.
\newblock \urlprefix\url{https://openreview.net/forum?id=rkgOLb-0W}.

\bibitem[{Shieber et~al.(1995)Shieber, Schabes \&
  Pereira}]{shieber_principles_1995}
Shieber, Stuart~M., Yves Schabes \& Fernando C.~N. Pereira. 1995.
\newblock Principles and implementation of deductive parsing.
\newblock \emph{The Journal of Logic Programming} 24(1). 3--36.
\newblock \doi{10.1016/0743-1066(95)00035-I}.

\bibitem[{Steedman(2000)}]{steedman_syntactic_2000}
Steedman, Mark. 2000.
\newblock \emph{The {Syntactic} {Process}}, vol.~24.
\newblock Cambridge, MA: MIT press.

\bibitem[{Stowell(1982)}]{stowell_tense_1982}
Stowell, Tim. 1982.
\newblock The tense of infinitives.
\newblock \emph{Linguistic Inquiry} 13(3). 561--570.

\bibitem[{Tai et~al.(2015)Tai, Socher \& Manning}]{tai_improved_2015}
Tai, Kai~Sheng, Richard Socher \& Christopher~D Manning. 2015.
\newblock Improved semantic representations from tree-structured long
  short-term memory networks.
\newblock In \emph{Proceedings of the 53rd {Annual} {Meeting} of the
  {Association} for {Computational} {Linguistics} and the 7th {International}
  {Joint} {Conference} on {Natural} {Language} {Processing}}, 1556--1566.
  Beijing, China: Association for Computational Linguistics.

\bibitem[{Uegaki(2015)}]{uegaki_interpreting_2015}
Uegaki, Wataru. 2015.
\newblock \emph{Interpreting questions under attitudes}.
\newblock Cambridge, MA: Massachusetts Institute of Technology {PhD} {Thesis}.

\bibitem[{White(accepted)}]{white_believing_accepted}
White, Aaron~Steven. accepted.
\newblock Believing and hoping whether.
\newblock \emph{Semantics and Pragmatics} .

\bibitem[{White \& Rawlins(2016)}]{white_computational_2016}
White, Aaron~Steven \& Kyle Rawlins. 2016.
\newblock A computational model of {S}-selection.
\newblock \emph{Semantics and Linguistic Theory} 26(0). 641--663.
\newblock \doi{10.3765/salt.v26i0.3819}.

\bibitem[{White \& Rawlins(2018)}]{white_role_2018}
White, Aaron~Steven \& Kyle Rawlins. 2018.
\newblock The role of veridicality and factivity in clause selection.
\newblock In Sherry Hucklebridge \& Max Nelson (eds.), \emph{Proceedings of the
  48th {Annual} {Meeting} of the {North} {East} {Linguistic} {Society}},
  221--234. Amherst, MA: GLSA Publications.

\bibitem[{White \& Rawlins(accepted)}]{white_frequency_accepted}
White, Aaron~Steven \& Kyle Rawlins. accepted.
\newblock Frequency, acceptability, and selection: A case study of clause
  embedding.
\newblock \emph{Glossa} .

\bibitem[{White et~al.(2018)White, Rudinger, Rawlins \&
  Van~Durme}]{white_lexicosyntactic_2018}
White, Aaron~Steven, Rachel Rudinger, Kyle Rawlins \& Benjamin Van~Durme. 2018.
\newblock Lexicosyntactic {Inference} in {Neural} {Models}.
\newblock In \emph{Proceedings of the 2018 {Conference} on {Empirical}
  {Methods} in {Natural} {Language} {Processing}}, 4717--4724. Brussels,
  Belgium: Association for Computational Linguistics.
\newblock \doi{10.18653/v1/D18-1501}.

\bibitem[{Williams et~al.(2018)Williams, Drozdov \&
  Bowman}]{williams_latent_2018}
Williams, Adina, Andrew Drozdov \& Samuel~R. Bowman. 2018.
\newblock Do latent tree learning models identify meaningful structure in
  sentences?
\newblock \emph{Transactions of the Association for Computational Linguistics}
  6. 253--267.
\newblock \doi{10.1162/tacl_a_00019}.

\bibitem[{Williamson(2019)}]{williamson_temporal_2019}
Williamson, Gregor. 2019.
\newblock The temporal orientation of infinitives.
\newblock In \emph{Proceedings of {Sinn} und {Bedeutung}}, vol.~23, 461--478.
\newblock Issue: 2.

\bibitem[{Wurmbrand(2014)}]{wurmbrand_tense_2014}
Wurmbrand, Susi. 2014.
\newblock Tense and aspect in {English} infinitives.
\newblock \emph{Linguistic Inquiry} 45(3). 403--447.

\bibitem[{Younger(1967)}]{younger_recognition_1967}
Younger, Daniel~H. 1967.
\newblock Recognition and parsing of context-free languages in time $n^3$.
\newblock \emph{Information and Control} 10(2). 189--208.

\bibitem[{Zettlemoyer \& Collins(2007)}]{zettlemoyer_online_2007}
Zettlemoyer, Luke \& Michael Collins. 2007.
\newblock Online learning of relaxed {CCG} grammars for parsing to logical
  form.
\newblock In \emph{Proceedings of the 2007 {Joint} {Conference} on {Empirical}
  {Methods} in {Natural} {Language} {Processing} and {Computational} {Natural}
  {Language} {Learning} ({EMNLP}-{C}o{NLL})}, 678--687. Prague, Czech Republic:
  Association for Computational Linguistics.
\newblock \urlprefix\url{https://www.aclweb.org/anthology/D07-1071}.

\bibitem[{Zettlemoyer \& Collins(2009)}]{zettlemoyer_learning_2009}
Zettlemoyer, Luke \& Michael Collins. 2009.
\newblock Learning context-dependent mappings from sentences to logical form.
\newblock In \emph{Proceedings of the {Joint} {Conference} of the 47th {Annual}
  {Meeting} of the {ACL} and the 4th {International} {Joint} {Conference} on
  {Natural} {Language} {Processing} of the {AFNLP}}, 976--984. Suntec,
  Singapore: Association for Computational Linguistics.
\newblock \urlprefix\url{https://www.aclweb.org/anthology/P09-1110}.

\bibitem[{Zettlemoyer \& Collins(2005)}]{zettlemoyer_learning_2005}
Zettlemoyer, Luke~S. \& Michael Collins. 2005.
\newblock Learning to map sentences to logical form: structured classification
  with probabilistic categorial grammars.
\newblock In \emph{Proceedings of the {Twenty}-{First} {Conference} on
  {Uncertainty} in {Artificial} {Intelligence}} {UAI}'05, 658--666. Arlington,
  Virginia, USA: AUAI Press.

\end{thebibliography}


\begin{addresses}
  \begin{address}
    Gene Louis Kim \\
    Department of Computer Science \\
    University of Rochester\\
    500 Joseph C. Wilson Blvd.\\
    Rochester, NY, USA 14627\\
    \email{gkim21@cs.rochester.edu}
  \end{address}
  \begin{address}
    Aaron Steven White\\ 
    Department of Linguistics\\
    University of Rochester\\
    500 Joseph C. Wilson Blvd.\\
    Rochester, NY, USA 14627\\
    \email{aaron.white@rochester.edu}
  \end{address}
\end{addresses}


\end{document}